%% file: arxiv.tex
\documentclass{article} 
\usepackage{iclr2027_conference,times}

\input{math_commands.tex}

\usepackage{hyperref}
\usepackage{url}
\usepackage{graphicx}
\usepackage{booktabs}
\usepackage{wrapfig}
\usepackage{float}
\usepackage{placeins}
\hypersetup{hidelinks}
\usepackage[table]{xcolor}
\usepackage[breakable]{tcolorbox}
\definecolor{oursblue}{HTML}{EEF3FA}
\usepackage{xspace}
\newcommand{\method}{\textsc{FrameMorrow}\xspace}

\title{FrameMorrow: Future-guided Frame Selection with Prospective Tokens for Long-Horizon Video Generation}

\author{
Bo Yin$^{1}$,
Xiaobin Hu$^{1}$,
Jiaqi Zhao$^{1, 2}$,
Shuicheng Yan$^{1}$,
\\[4pt]
$^{1}$National University of Singapore \\
$^{2}$Harbin Institute of Technology (Shenzhen)
}

\iclrfinalcopy 
\begin{document}

\maketitle

\begin{abstract}
Long-horizon video generation requires models to effectively leverage an increasingly long generation history.
As the generated history grows, retaining all previous content becomes increasingly expensive and redundant, making effective historical selection essential. 
Existing approaches often determine historical relevance from the current content. 
However, information that is relevant to the present is not necessarily useful for future generation, while seemingly less relevant history may become important later.
Our key insight is that \emph{historical information should be selected according to its relevance to future information needs}. Capturing these needs does not require generating the full future.
A compact representation of what becomes important next is sufficient to guide historical selection. 
Building on this insight, we propose \textbf{\method}, a prospective frame selector that predicts a small set of \emph{prospective tokens} representing future information needs and uses them to identify relevant information from history.
\method selects explicit historical frames rather than model-specific internal states, enabling plug-and-play integration across diverse generators, including closed-source models, with little additional inference cost.
We evaluate \method across five benchmarks and 11 generative models spanning long-video generation, interactive generation, and action-conditioned world models. 
Extensive experiments demonstrate improvements in long-range consistency, visual quality, and action alignment across diverse generation settings. The project page is at \url{https://yinbo0927.github.io/FrameMorrow/}.
\end{abstract}

\input{sec/introdutcion}
\input{sec/related_work}

\input{sec/method}
\input{sec/experiment}

\input{sec/conclusion}

\bibliography{iclr2027_conference}
\bibliographystyle{iclr2027_conference}

\newpage
\appendix
\input{sec/appendix}

\end{document}

%% file: math_commands.tex
\usepackage{amsmath,amsfonts,bm}

\def\eqref#1{equation~\ref{#1}}

\def\1{\bm{1}}

\DeclareMathAlphabet{\mathsfit}{\encodingdefault}{\sfdefault}{m}{sl}
\SetMathAlphabet{\mathsfit}{bold}{\encodingdefault}{\sfdefault}{bx}{n}



%% file: sec/introdutcion.tex
\section{Introduction}
\label{sec:introduction}

Long-horizon video generation requires models not only to continuously extend visual content, but also to preserve coherent subjects, objects, and scenes over time~\citep{yang2025longlive,zhang2025storymem, yu2026dual, yin2026spot, zhao2026quantwm}. As generation proceeds, visual details generated earlier in the video gradually fall outside the model's limited input window~\citep{hu2026longlive}. When such content becomes relevant again, subjects can change appearance, object states can become inconsistent, and revisited scenes can no longer match their earlier observations~\citep{xiao2026worldmem}. Keeping all historical information is impractical, as it increases computation and introduces a large amount of redundant context~\citep{yi2025deep,ji2025memflow, nie2026skillgraph}. Moreover, not all historical information is equally useful at every generation step~\citep{ji2025memflow,an2026onestory, yin2025fera}. The key challenge is therefore to selectively preserve the historical information that matters for future generation.

\begin{figure}[t]
    \centering
    \includegraphics[width=\linewidth]{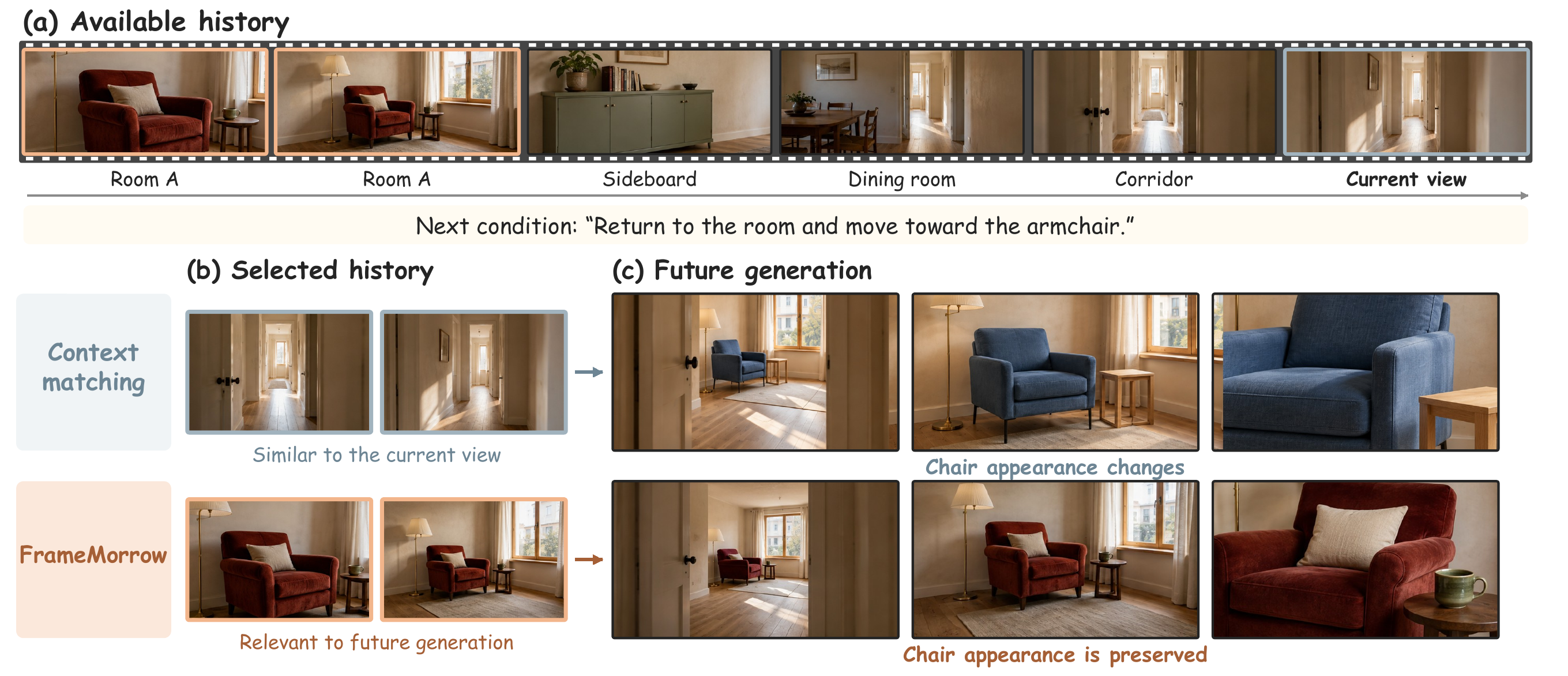}
    \vspace{-7mm}
    \caption{\textbf{Motivation for future-relevant history selection.}
    Given the same available history and known next condition, current-context
    matching favors observations similar to the current view, whereas \method
    selects earlier history that is more relevant to the upcoming generation.
    The selected history helps preserve previously observed visual details in
    the generated continuation.}
    \label{fig:teaser}
\end{figure}

Many existing methods identify useful history based on its relevance to the current content~\citep{zhang2025egolcd}, often using the recent visual context to retrieve related information from the past~\citep{hu2026longlive,ye2026dysink,wang2026dual,ding2026layerrecall}. Such a strategy mainly answers which historical information is most relevant to what is visible now. However, this can differ from what will actually be useful for future generation. Information that closely matches the present may provide little additional value for what comes next, while earlier information that appears less relevant now can become important again as the generation evolves. As illustrated in Fig.~\ref{fig:teaser}, matching the current view can favor
visually similar history, while information that is less relevant to the
present may better support the upcoming generation. Therefore, the importance of historical information should not be determined only by its relevance to the present, but also by its potential usefulness for the future. \emph{Can historical information be selected according to its relevance to the future rather than to the current content?}

However, the challenge is that the future has not been generated when the selection is made. Since the goal is to determine which historical information will be useful for future generation, there is no need to generate the full future itself. Instead, it is sufficient to predict a compact representation that captures future information needs and serves as a proxy for the future. Such a representation can then guide historical selection toward information that is likely to matter next.

Motivated by this, we propose \method, which predicts a small set of \emph{prospective tokens} as compact representations of future information needs. These tokens capture what potentially become important next and are used to identify historical information that is relevant to future generation. \method then selects explicit historical frames based on this relevance. Because it selects explicit historical frames rather than model-specific internal states, the same selector can be used across different generators, while each model processes the selected frames in its native way. This makes \method plug-and-play across different generators, including closed-source models.
Moreover, predicting only a few prospective tokens and selecting only a few historical frames keeps \method lightweight with little additional inference cost.

Our contributions are as follows:
\begin{itemize}
    \item We formulate using future information needs as the condition for frame selection. Selection based only on current content cannot predict future information needs and retrieve frames that are relevant to the present but unhelpful for future generation.

    \item We propose \method, which introduces \emph{prospective tokens} to explicitly represent future information needs and identify relevant information from the history.
    
    \item We design \method as a \emph{plug-and-play}, lightweight selector that outputs explicit historical frames, enabling broad compatibility across different generators, including closed-source models, with little inference overhead.

    \item We extensively evaluate \method across five benchmarks and 11 generative models, covering long-video generation, interactive generation, and action-conditioned world models, with improvements in long-range consistency, visual quality, and action alignment.
\end{itemize}

%% file: sec/related_work.tex
\section{Related Work}
\label{sec:related_work}

\noindent\textbf{Long-Horizon Video Generation.}
Autoregressive video models extend videos by conditioning new content on
previously generated outputs~\citep{yin2025slow,huang2026self}.
CausVid~\citep{yin2025slow} distills a bidirectional diffusion model into a few-step causal
generator for streaming synthesis.
Self-Forcing~\citep{huang2026self} reduces the training--inference gap by using self-generated
context during training.
LongLive~\citep{yang2025longlive} supports real-time long-video generation with changing text
prompts, while Matrix-Game 3.0~\citep{wang2026matrix} extends interactive
world generation with action control and long-horizon
memory.
Rather than developing another generation backbone, our work provides
a plug-and-play frame selector that supports diverse existing generators
by supplying relevant historical information.

\noindent\textbf{Historical Information Selection.}
Historical information can be retrieved as visual context or retained
within a generator's internal cache.
LongLive-RAG retrieves historical latents using the latest generated
content~\citep{hu2026longlive}, while DySink selects visually relevant
historical frames as dynamic frame sinks~\citep{ye2026dysink}.
For narrative generation, MemFlow retrieves history using the upcoming
chunk prompt~\citep{ji2025memflow}, and Memento separately retrieves
identity evidence and short-range shot cues~\citep{wei2026memento}.
Mem-World combines planned actions with scene geometry to retrieve
relevant historical observations~\citep{zheng2026mem}.
For cache management, PaFu-KV learns token salience from a bidirectional
teacher~\citep{chen2026past}, while Future Forcing constructs future-query
proxies to guide cache eviction and merging~\citep{luo2026future}.
Unlike existing approaches that determine historical relevance from
observed content or model-specific memory states, our method selects
history according to its relevance to future information needs.

\noindent\textbf{Learning to Select Frames.}
Learned frame selection reduces redundant visual input for video
understanding.
Frame-Voyager learns query-conditioned frame combinations from rankings
provided by a video-language model~\citep{yu2025frame}.
FrameOracle predicts both query-relevant frames and an adaptive frame
budget~\citep{li2025frameoracle}.
Related approaches learn selection through multimodal-model
supervision~\citep{hu2025m}, flexible selection
policies~\citep{buch2025flexible}, or reinforcement-learning
rewards~\citep{qin2026efficient}.
These methods select evidence to answer questions or reason about
an available video.
Unlike frame selectors designed for fully observed videos, \method
predicts \emph{prospective tokens} to represent information needs for
content that has not yet been generated, enabling lightweight selection
of relevant historical frames.


%% file: sec/method.tex
\section{Method}
\label{sec:method}

\begin{figure}[t]
    \centering
    \includegraphics[width=\textwidth]{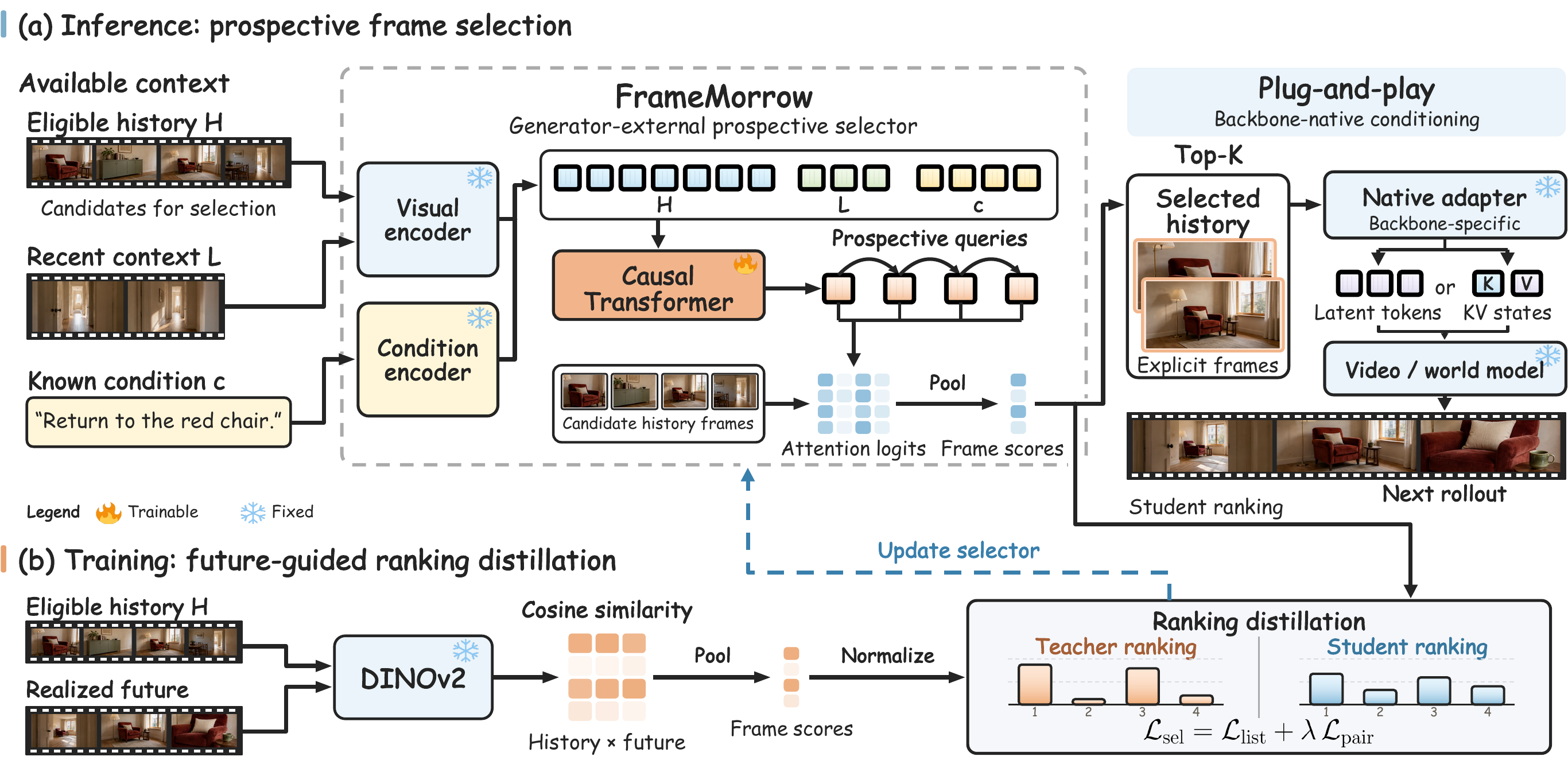}
    \vspace{-5mm}
    \caption{\textbf{Overview of \method.}
    (a) At inference, the selector autoregressively predicts prospective
    tokens from history, recent context, and the rollout condition.
    These tokens score eligible historical frames and guide top-$K$
    selection. Each frozen generator processes the selected frames through
    its own conditioning mechanism.
    (b) During training, frozen DINOv2 features provide future-grounded
    frame rankings, which supervise the prospective tokens through
    listwise and pairwise ranking losses.}
    \label{fig:overview}
\end{figure}

\noindent
\textbf{Overview.}
\method predicts a small set of \emph{prospective tokens} to represent
future information needs by given eligible history $\mathcal H_t$,
recent context $\mathcal L_t$, and the known rollout condition $c_t^+$.
These tokens score historical relevance and guide frame selection.
During training, a frozen visual teacher ranks historical frames by
their correspondence with future content, providing ranking supervision
for the tokens.
At inference, \method outputs explicit historical frames, which augment
a compatible frozen generator through its own conditioning mechanism
(Fig.~\ref{fig:overview}).

\subsection{Prospective Frame Selection}
\label{sec:problem}

At rollout step $t$, we distinguish the backbone's recent context from the
eligible long-term history:
\begin{equation}
    \mathcal L_t=\{x_{t-L+1},\ldots,x_t\},
    \qquad
    \mathcal H_t=\{x_{\tau_i}\}_{i=1}^{N},
    \quad \tau_i\leq t-L.
\end{equation}
Both sets are observed by the selector, but only $\mathcal H_t$ is eligible for
retrieval. Recent context thus informs which past evidence to recall without
occupying the long-term memory budget.
The condition $c_t^+$ is a text prompt, action sequence, or control signal
available at step $t$. It contains no later user input, environment feedback,
or future observation. For a memory budget $K$, we select
\begin{equation}
    \hat{\mathbf r}_t=S_{\theta}(\mathcal H_t,\mathcal L_t,c_t^+)\in\mathbb R^N,
    \qquad
    \mathcal I_t=\operatorname{TopK}(\hat{\mathbf r}_t,K),
    \qquad
    \mathcal M_t=\mathcal H_t[\mathcal I_t].
\end{equation}
The selected frames are ordered by their timestamps before being passed to
the backbone.

\subsection{Prospective Tokens}
\label{sec:psq}

To represent future information needs without generating future content,
a compact causal Transformer predicts prospective tokens from the
observed history, recent context, and rollout condition. A frozen visual encoder $E_v$ and learned projection $P_v$ encode
each observed frame as $\mathbf z(x)=P_v[E_v(x)]\in\mathbb R^d$.
We form the temporally ordered sequences
$\mathbf H_t=[\mathbf h_1,\ldots,\mathbf h_N]$, with
$\mathbf h_i=\mathbf z(x_{\tau_i})$, and
$\mathbf L_t=[\mathbf z(x_{t-L+1}),\ldots,\mathbf z(x_t)]$.
A frozen modality-specific encoder $E_c$ and learned projection $P_c$ produce
$\mathbf C_t^+=P_c[E_c(c_t^+)]$.
Selector weights are shared across compatible backbones within each condition
modality. Different modalities share the architecture and interface.

After processing $[\mathbf H_t,\mathbf L_t,\mathbf C_t^+]$, the selector
autoregressively predicts $M$ prospective tokens:
\begin{equation}
    \mathbf q_t^m=F_{\psi}(\mathbf H_t,\mathbf L_t,\mathbf C_t^+,\mathbf q_t^{<m}),
    \qquad m=1,\ldots,M.
\end{equation}
All predicted tokens are retained as
$\mathbf Q_t=[\mathbf q_t^1,\ldots,\mathbf q_t^M]\in\mathbb R^{M\times d}$
($M=4$ by default).
Each token can condition on preceding tokens, allowing their predictions
to depend on one another.
Their prospective role is learned through future-grounded ranking
supervision, without assigning predefined future factors to individual
tokens.
At each rollout step, the tokens are regenerated from the currently
available inputs.
With $\theta=\{P_v,P_c,\psi,W_Q,W_K\}$, the prospective tokens serve as
attention queries over eligible historical frames.
We compute scaled dot-product attention logits and aggregate them
with a smooth maximum:
\begin{equation}
    a_{t,m,i}=\frac{(W_Q\mathbf q_t^m)^\top(W_K\mathbf h_i)}{\sqrt d},
    \qquad
    \hat r_{t,i}=\tau_q\log\!\left(\frac{1}{M}\sum_{m=1}^{M}e^{a_{t,m,i}/\tau_q}\right).
\end{equation}
Here $\tau_q>0$ controls aggregation sharpness.
A frame can score highly by matching any prospective token. The objective supervises the aggregated ranking without explicitly enforcing diversity among tokens or selected frames.

\subsection{Future-Grounded Ranking Distillation}
\label{sec:future_supervision}

We train the prospective tokens by matching their predicted frame rankings
to rankings derived from future content.
Each training trajectory supplies eligible history, recent context, a known
condition, and a realized continuation
$\mathcal Y_t^+=\{x_{t+j}\}_{j=1}^{H}$.
The teacher compares candidate and continuation frames using a frozen DINOv2
encoder $\Phi$. Historical views of subjects or scenes that recur in the
continuation can provide reusable visual evidence, motivating the
correspondence target
\begin{equation}
    R_{t,i,j}^{F}=\operatorname{cos}(\Phi(x_{\tau_i}),\Phi(x_{t+j})),
    \qquad
    r_{t,i}^{F}=\tau_f\log\!\left(\frac{1}{H}\sum_{j=1}^{H}e^{R_{t,i,j}^{F}/\tau_f}\right).
\end{equation}
This smooth maximum favors correspondence with at least part of the
continuation. It provides a visual relevance proxy, rather than a measurement
of incremental generation utility beyond the recent context or a guarantee of
coverage across the continuation. We empirically examine the relationship between this proxy and downstream
generation utility in Appendix~\ref{app:utility_validation}. The teacher uses only
$(\mathcal H_t,\mathcal Y_t^+)$, while the student receives only
$(\mathcal H_t,\mathcal L_t,c_t^+)$ during both training and inference.

We collect $\mathbf r_t^F=[r_{t,1}^F,\ldots,r_{t,N}^F]$ and normalize teacher
and student scores into ranking distributions:
\begin{equation}
    \mathbf p_t^F=\operatorname{Softmax}(\mathbf r_t^F/\tau_r),
    \qquad
    \hat{\mathbf p}_t=\operatorname{Softmax}(\hat{\mathbf r}_t/\tau_s).
\end{equation}
The temperatures $\tau_f,\tau_r,\tau_s$ are positive.
We retain pairwise orderings separated by a margin $\delta>0$,
$\mathcal P_t=\{(i,j)\mid r_{t,i}^F\geq r_{t,j}^F+\delta\}$,
and optimize
\begin{equation}
    \mathcal L_{\mathrm{sel}}=\mathcal L_{\mathrm{list}}+\lambda\mathcal L_{\mathrm{pair}},
\end{equation}
where the listwise term aligns the full distribution and the pairwise term
preserves orderings separated by the teacher-score margin:
\begin{equation}
    \mathcal L_{\mathrm{list}}=-\sum_{i=1}^{N}p_{t,i}^F\log\hat p_{t,i},
    \qquad
    \mathcal L_{\mathrm{pair}}=\frac{1}{|\mathcal P_t|}\sum_{(i,j)\in\mathcal P_t}
    \operatorname{softplus}[-(\hat r_{t,i}-\hat r_{t,j})].
\end{equation}
We set $\mathcal L_{\mathrm{pair}}=0$ when $\mathcal P_t$ is empty.
At inference, continuation-based target construction is removed. The selector
still encodes observed frames with $E_v$ and predicts prospective tokens from
$(\mathcal H_t,\mathcal L_t,c_t^+)$, without accessing $\mathcal Y_t^+$.

\subsection{Plug-and-Play Integration}
\label{sec:plugin}

\method outputs explicit historical frames, while each generator processes
these frames through its own conditioning mechanism.
For a frozen backbone $b$ that supports reference-frame or memory
conditioning, its native adapter $\Gamma_b$ maps the selected frames
to the representation already accepted by that model:
\begin{equation}
    \mathbf M_t^{(b)}=\Gamma_b(\mathcal M_t),
    \qquad
    \hat{\mathcal Y}_t^+=G_{\omega_b}(\mathcal L_t,c_t^+,\mathbf M_t^{(b)}),
    \qquad \omega_b\ \text{is frozen}.
\end{equation}
For example, a reference-frame interface applies its existing image
preprocessing and reference encoder to $\mathcal M_t$.
No new conditioning module is trained.
The selector decides which historical frames to provide, while each
generator retains its own way of processing the selected content.

Predicting a small number of prospective tokens avoids generating an
additional future video for selection.
A fixed $K$ bounds the number of additional frames supplied to the
generator.
Selector processing still depends on the candidate count,
recent-context length, and condition length.

%% file: sec/experiment.tex
\begin{table}[t]
    \centering
    \setlength{\abovecaptionskip}{0pt}
    \setlength{\belowcaptionskip}{4pt}
    \caption{\textbf{VBench-Long results for 60-second generation on all 128 MovieGenBench prompts.}
    Each block compares long-context methods on one backbone.
    Average rank is computed over six metrics. Bold and underline mark
    the best and second-best results within each block.}
    \label{tab:moviegenbench}
    \setlength{\tabcolsep}{3.2pt}
    \resizebox{\linewidth}{!}{%
    \begin{tabular}{lccccccc}
        \toprule
        \textbf{Method}
        & \textbf{Subject $\uparrow$}
        & \textbf{Background $\uparrow$}
        & \textbf{Motion $\uparrow$}
        & \textbf{Dynamic $\uparrow$}
        & \textbf{Aesthetic $\uparrow$}
        & \textbf{Imaging $\uparrow$}
        & \textbf{Avg. Rank $\downarrow$} \\
        \midrule

        \rowcolor{gray!10}
        Self-Forcing
        & 95.84
        & 95.27
        & 98.20
        & \underline{51.72}
        & 56.05
        & 62.22
        & 4.33 \\

        \hspace{0.5em}$+\infty$-RoPE
        & 97.24
        & 96.24
        & 98.58
        & 46.64
        & 56.09
        & 63.28
        & 3.17 \\

        \hspace{0.5em}+Deep Forcing
        & 96.08
        & 95.38
        & 98.24
        & 41.44
        & 56.68
        & 60.81
        & 4.17 \\

        \hspace{0.5em}+LongLive-RAG
        & \underline{97.60}
        & \textbf{96.51}
        & \textbf{98.70}
        & 44.69
        & \underline{57.19}
        & \underline{64.97}
        & \underline{2.00} \\

        \rowcolor{oursblue}
        \hspace{0.5em}\textbf{+\method (Ours)}
        & \textbf{97.71}
        & \underline{96.45}
        & \underline{98.60}
        & \textbf{64.56}
        & \textbf{57.27}
        & \textbf{68.47}
        & \textbf{1.33} \\

        \midrule

        \rowcolor{gray!10}
        LongLive 1.0
        & 97.13
        & 95.89
        & 98.61
        & 44.56
        & 58.17
        & 67.56
        & 3.83 \\

        \hspace{0.5em}$+\infty$-RoPE
        & 97.00
        & 95.85
        & 98.53
        & \textbf{53.36}
        & 57.48
        & 66.94
        & 4.25 \\

        \hspace{0.5em}+Deep Forcing
        & \underline{97.17}
        & 96.04
        & \underline{98.73}
        & 45.13
        & 57.48
        & 67.27
        & 3.42 \\

        \hspace{0.5em}+LongLive-RAG
        & \textbf{97.32}
        & \underline{96.08}
        & 98.62
        & 49.90
        & \underline{58.30}
        & \underline{67.79}
        & \underline{2.25} \\

        \rowcolor{oursblue}
        \hspace{0.5em}\textbf{+\method (Ours)}
        & \textbf{97.32}
        & \textbf{96.17}
        & \textbf{98.75}
        & \underline{50.42}
        & \textbf{58.75}
        & \textbf{67.82}
        & \textbf{1.25} \\

        \midrule

        \rowcolor{gray!10}
        Causal Forcing
        & 93.52
        & 94.12
        & 95.74
        & 72.32
        & 51.24
        & 62.30
        & 4.83 \\

        \hspace{0.5em}$+\infty$-RoPE
        & 93.81
        & 93.78
        & 96.09
        & \textbf{92.47}
        & 54.42
        & 67.50
        & 3.33 \\

        \hspace{0.5em}+Deep Forcing
        & 94.27
        & 94.18
        & \textbf{96.62}
        & 78.59
        & 52.12
        & 64.25
        & 3.17 \\

        \hspace{0.5em}+LongLive-RAG
        & \underline{94.29}
        & \underline{94.24}
        & 96.48
        & 88.20
        & \underline{54.95}
        & \underline{68.16}
        & \underline{2.33} \\

        \rowcolor{oursblue}
        \hspace{0.5em}\textbf{+\method (Ours)}
        & \textbf{94.54}
        & \textbf{94.28}
        & \underline{96.59}
        & \underline{89.74}
        & \textbf{56.27}
        & \textbf{68.76}
        & \textbf{1.33} \\

        \bottomrule
    \end{tabular}}
\end{table}

\section{Experiments}
\label{sec:experiments}

We evaluate \method on five benchmarks covering long-video generation,
single-shot interactive generation, multi-shot interactive generation,
closed-source generation, and action-conditioned world models.

\subsection{Experimental Setup}
\label{sec:exp_setup}

\noindent
\textbf{Backbones and baselines.}
For long-video generation, we evaluate Self-Forcing~\citep{huang2026self},
LongLive 1.0~\citep{yang2025longlive}, and Causal Forcing~\citep{zhu2026causal}.
Long-context baselines include $\infty$-RoPE~\citep{yesiltepe2026infinity}, Deep Forcing~\citep{yi2025deep},
and LongLive-RAG~\citep{hu2026longlive}. Interactive video experiments
additionally include CausVid~\citep{yin2025slow}, LongLive 2.0~\citep{chen2026longlive}, and ShotStream~\citep{luo2026shotstream}.
World-model experiments use Matrix-Game 3.0~\citep{wang2026matrix}, WorldMem~\citep{xiao2026worldmem},
and YuMe 1.5~\citep{mao2026yume1}. Closed-source model experiments use Seedance 2.0 and Kling O3 through
their public reference-conditioning interfaces.

\noindent
\textbf{Offline training and integration.}
For text-conditioned generation, we train on 10K OpenVidHD~\cite{nan2025openvid} videos organized into streaming prompts.
For action-conditioned world models, we use 10K chunk-aligned examples from Sekai Game-Walking with pseudo-actions derived from camera motion.
In both settings, realized future observations are used only to construct future-grounded ranking supervision.
At inference, selected frames are passed through each frozen backbone's existing conditioning interface.
Unless otherwise stated, we use $K=4$ historical frames and $M=4$ prospective tokens while retaining each backbone's native recent context.
Further details are provided in Appendix~\ref{app:implementation}.

\noindent
\textbf{Evaluation metrics.}
For MovieGenBench, we report the six VBench-Long dimensions and average rank
across these dimensions within each backbone. Ties receive their average rank.
For interactive long video generation, we report overall quality, consistency,
and aesthetic scores. We measure semantic adherence using CLIP scores between
each 10-second segment and its corresponding prompt. For closed-source generation, we report the same overall quality, consistency, and aesthetic scores as in interactive video generation. For interactive world models, we report visual quality,
temporal quality, and action alignment. Qwen3-VL-8B-Instruct evaluates action alignment
by assessing whether generated rollouts follow the supplied actions.

\begin{table}[t]
    \centering
    \setlength{\abovecaptionskip}{0pt}
    \setlength{\belowcaptionskip}{4pt}
    \caption{\textbf{Interactive video generation over 60 seconds.}
    Single-shot and multi-shot results are grouped separately.
    Bold marks the better result within each backbone pair.}
    \label{tab:interactive_long_video}

    \setlength{\tabcolsep}{2.8pt}
    \renewcommand{\arraystretch}{1.02}

    \resizebox{\linewidth}{!}{%
    \begin{tabular}{lcccccccccc}
        \toprule

        & \multicolumn{3}{c}{\textbf{Overall}}
        & \multicolumn{7}{c}{\textbf{Segment-wise CLIP Score $\uparrow$}} \\
        \cmidrule(lr){2-4}
        \cmidrule(lr){5-11}

        \textbf{Model}
        & \textbf{Quality $\uparrow$}
        & \textbf{Consistency $\uparrow$}
        & \textbf{Aesthetic $\uparrow$}
        & \textbf{0--10s}
        & \textbf{10--20s}
        & \textbf{20--30s}
        & \textbf{30--40s}
        & \textbf{40--50s}
        & \textbf{50--60s}
        & \textbf{Avg.} \\

        \midrule

        \rowcolor{gray!10}\multicolumn{11}{l}{\textbf{Single-shot}} \\
        \addlinespace[1.5pt]

        LongLive 1.0
        & 83.49
        & 92.62
        & 64.21
        & 30.71
        & 29.41
        & 27.35
        & 28.45
        & 27.63
        & 28.95
        & 28.75 \\

        \rowcolor{oursblue}
        \hspace{0.5em}+\method (Ours)
        & \textbf{84.06}
        & \textbf{93.21}
        & \textbf{64.56}
        & \textbf{30.72}
        & \textbf{29.65}
        & \textbf{28.11}
        & \textbf{28.78}
        & \textbf{28.03}
        & \textbf{29.18}
        & \textbf{29.08} \\

        \addlinespace[1pt]

        Self-Forcing
        & 78.71
        & 84.95
        & 58.16
        & 30.40
        & 29.52
        & 26.93
        & 25.10
        & 23.37
        & 23.02
        & 26.39 \\

        \rowcolor{oursblue}
        \hspace{0.5em}+\method (Ours)
        & \textbf{81.74}
        & \textbf{89.30}
        & \textbf{60.62}
        & \textbf{30.56}
        & \textbf{29.67}
        & \textbf{28.34}
        & \textbf{27.12}
        & \textbf{26.60}
        & \textbf{26.01}
        & \textbf{28.05} \\

        \addlinespace[1pt]

        Causal Forcing
        & 76.00
        & 80.46
        & 55.45
        & \textbf{29.92}
        & \textbf{27.37}
        & 23.89
        & 22.43
        & 21.02
        & 22.56
        & 24.53 \\

        \rowcolor{oursblue}
        \hspace{0.5em}+\method (Ours)
        & \textbf{78.66}
        & \textbf{83.48}
        & \textbf{58.75}
        & 29.88
        & 27.15
        & \textbf{24.27}
        & \textbf{23.76}
        & \textbf{23.30}
        & \textbf{23.08}
        & \textbf{25.24} \\

        \addlinespace[1pt]

        CausVid
        & 81.94
        & 89.77
        & 63.53
        & \textbf{29.92}
        & 29.00
        & 28.34
        & 28.22
        & 28.19
        & 28.08
        & 28.63 \\

        \rowcolor{oursblue}
        \hspace{0.5em}+\method (Ours)
        & \textbf{82.26}
        & \textbf{90.52}
        & \textbf{63.64}
        & 29.81
        & \textbf{29.21}
        & \textbf{28.86}
        & \textbf{28.82}
        & \textbf{28.81}
        & \textbf{28.64}
        & \textbf{29.03} \\

        \midrule

        \rowcolor{gray!10}\multicolumn{11}{l}{\textbf{Multi-shot}} \\
        \addlinespace[1.5pt]

        LongLive 2.0
        & 82.77
        & 93.14
        & 61.32
        & 29.27
        & \textbf{28.12}
        & \textbf{28.12}
        & 28.53
        & 28.04
        & 28.02
        & 28.35 \\

        \rowcolor{oursblue}
        \hspace{0.5em}+\method (Ours)
        & \textbf{83.18}
        & \textbf{93.49}
        & \textbf{61.68}
        & \textbf{29.28}
        & 28.05
        & 28.06
        & \textbf{28.71}
        & \textbf{28.29}
        & \textbf{28.24}
        & \textbf{28.44} \\

        \addlinespace[1pt]

        ShotStream
        & 83.12
        & 92.75
        & \textbf{62.30}
        & \textbf{30.15}
        & \textbf{28.87}
        & 28.00
        & 29.14
        & 29.19
        & 29.19
        & 29.09 \\

        \rowcolor{oursblue}
        \hspace{0.5em}+\method (Ours)
        & \textbf{84.85}
        & \textbf{96.76}
        & 61.74
        & 29.68
        & 28.58
        & \textbf{28.38}
        & \textbf{29.36}
        & \textbf{29.43}
        & \textbf{29.51}
        & \textbf{29.16} \\

        \bottomrule
    \end{tabular}}
\end{table}

\begin{figure}[!t]
    \setlength{\abovecaptionskip}{6pt}
    \setlength{\belowcaptionskip}{0pt}
    \centering
\includegraphics[width=\linewidth]{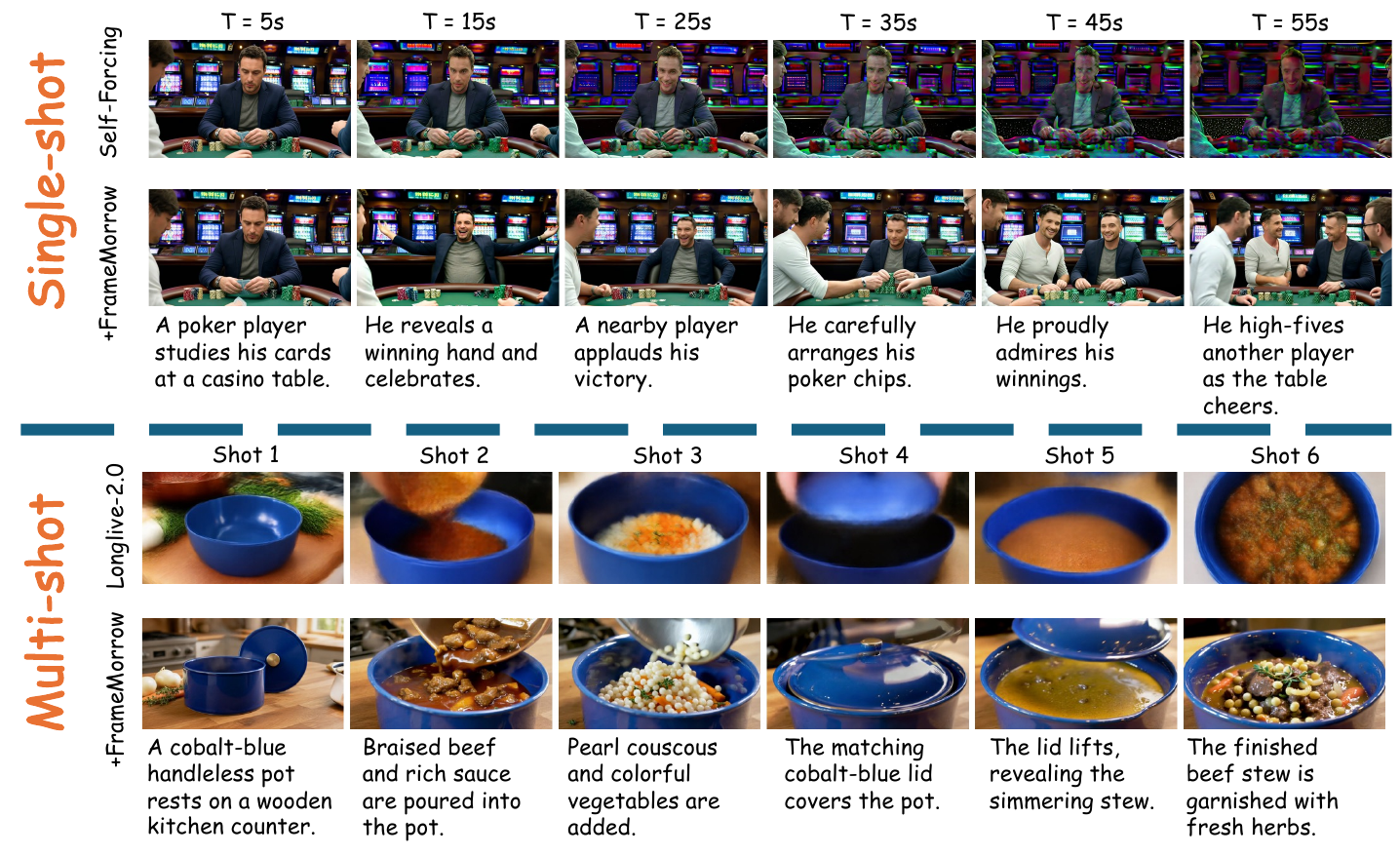}
 \vspace{-6mm}
    \caption{\textbf{Qualitative comparisons for interactive video generation.}
    Top: single-shot generation with Self-Forcing under sequential prompts.
    Bottom: multi-shot generation with LongLive 2.0.
    Each pair compares the backbone alone with its \method-augmented variant.}
\label{fig:interactive_qualitative}
\end{figure}

\subsection{Main Result}

\noindent
\textbf{Long video generation.}
\label{sec:long_video_results}
Table~\ref{tab:moviegenbench} compares long-context mechanisms for 60-second
MovieGenBench generation. \method achieves the best average rank on all three
backbones: 1.33 for Self-Forcing, 1.25 for LongLive 1.0, and 1.33 for Causal
Forcing. With Self-Forcing, it improves imaging quality from 62.22 to 68.47
and dynamic degree from 51.72 to 64.56. The advantage is not uniform across
metrics, with LongLive-RAG retaining higher background and motion scores on
Self-Forcing. The consistent average-rank advantage supports the effectiveness of frame
selection across these generators, with trade-offs in individual metrics.

\noindent
\textbf{Interactive video generation.}
\label{sec:interactive_long_video}
We evaluate interactive generation where the generation condition changes
over time, covering both single- and multi-shot settings.
Table~\ref{tab:interactive_long_video} reports improvements in overall
quality and consistency across both settings.

\begin{wraptable}{r}{0.48\linewidth}
    \centering
    \vspace{-5pt}
    \setlength{\abovecaptionskip}{0pt}
    \setlength{\belowcaptionskip}{3pt}

    \caption{\textbf{Closed-source generation.}
    All baselines use the same reference budget.}
    \label{tab:closed_source}

    \setlength{\tabcolsep}{3.0pt}
    \small

    \resizebox{\linewidth}{!}{%
    \begin{tabular}{lccc}
        \toprule
        \textbf{Model}
        & \textbf{Quality $\uparrow$}
        & \textbf{Consistency $\uparrow$}
        & \textbf{Aesthetic $\uparrow$} \\
        \midrule

        Seedance 2.0
        & 86.80 & 94.02 & 65.61 \\

        \hspace{0.5em}+ Uniform
        & 86.47 & 93.54 & 65.28 \\

        \rowcolor{oursblue}
        \hspace{0.5em}+\method
        & \textbf{87.46}
& \textbf{95.31}
& \textbf{65.92} \\

        \addlinespace[1.5pt]

        Kling O3
        & 83.47 & 91.28 & 63.41 \\

        \hspace{0.5em}+ Uniform
        & 83.21 & 90.84 & 63.18 \\

        \rowcolor{oursblue}
        \hspace{0.5em}+\method
        & \textbf{84.06}
& \textbf{92.61}
& \textbf{63.77} \\

        \bottomrule
    \end{tabular}}

    \vspace{-6pt}
\end{wraptable}

For
\emph{single-shot generation}, 
\method improves overall quality and consistency across all evaluated
backbones. For Self-Forcing, consistency rises from 84.95 to 89.30.
Its segment-wise CLIP-score gain increases from 0.16 in the first 10 seconds
to 2.99 in the last 10 seconds, indicating larger improvements in prompt
alignment at later stages of generation.
Figure~\ref{fig:interactive_qualitative} shows a representative example,
where the native Self-Forcing sequence develops strong color artifacts,
while \method depicts the successive actions at the poker table more clearly.

\begin{wrapfigure}{r}{0.48\linewidth}
    \centering
    \vspace{-2pt}
    \vspace{-5mm}
    \includegraphics[width=\linewidth]{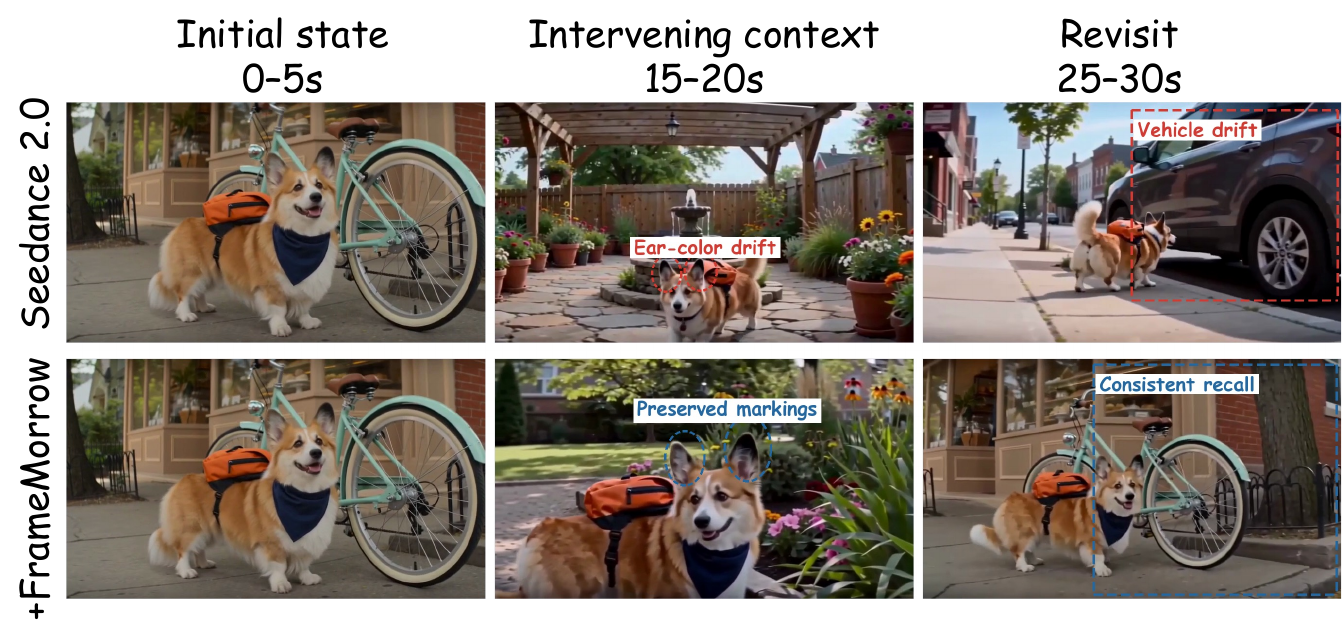}
    \vspace{-7mm}
    \caption{\textbf{Seedance 2.0 revisit case.}}
    \label{fig:closed_qual}

    \vspace{-6pt}
\end{wrapfigure}

For
\emph{multi-shot generation}, 
\method also improves overall quality and consistency for both evaluated
backbones. In particular, ShotStream gains 4.01 consistency points, although
its aesthetic score decreases by 0.56.
In the qualitative example, \method better preserves the blue pot and the
specified ingredients across multiple successive shot changes.

\noindent
\textbf{Extension to closed-source generators.}
To test whether \method remains effective without access to generator
internals, we further evaluate it on Seedance 2.0 and Kling O3 through their
public reference-conditioning interfaces. For all variants, the latest
generated clip is provided through the native video-reference interface.
Uniform-$K$ and \method additionally receive the same number of historical
frames through the image-reference interface, differing only in how these
frames are selected. Table~\ref{tab:closed_source} reports 30 cases 30-second
multi-turn generation results using the same overall metrics as our
open-source evaluation. Compared with the original generators, \method improves consistency
by 1.29 points on Seedance 2.0 and 1.33 points on Kling O3.
It also outperforms uniform selection on all three reported metrics
for both models.
These results support plug-and-play use with closed-source generators
and show the benefit of selecting relevant references under a fixed
reference budget. Figure~\ref{fig:closed_qual} shows a representative
revisit case on Seedance 2.0. While the original model exhibits subject and
object drift after intervening interactions, \method better preserves both
the established subject appearance and the previously observed vehicle.

\begin{table}[t]
    \centering
    \setlength{\abovecaptionskip}{0pt}
    \setlength{\belowcaptionskip}{4pt}
    \caption{\textbf{Action-conditioned interactive world generation.}
    Subject, background, anti-flicker, motion, and action-alignment scores
    lie in $[0,1]$. Imaging quality is on a $0$--$100$ scale.
    Shaded rows denote \method, and bold indicates the better result
    within each base-model pair.}
    \label{tab:world_models}

    \setlength{\tabcolsep}{3.5pt}
    \renewcommand{\arraystretch}{1.02}

    \resizebox{\linewidth}{!}{%
    \begin{tabular}{lcccccc}
        \toprule

        & \multicolumn{3}{c}{\textbf{Visual Quality}}
        & \multicolumn{2}{c}{\textbf{Temporal Quality}}
        & \multicolumn{1}{c}{\textbf{Interaction}} \\
        \cmidrule(lr){2-4}
        \cmidrule(lr){5-6}
        \cmidrule(lr){7-7}

        \textbf{Model}
        & \textbf{Subject $\uparrow$}
        & \textbf{Background $\uparrow$}
        & \textbf{Imaging $\uparrow$}
        & \textbf{Anti-flicker $\uparrow$}
        & \textbf{Motion $\uparrow$}
        & \textbf{Action Alignment $\uparrow$} \\

        \midrule

        Matrix-Game 3.0
        & 0.801
        & 0.894
        & 68.20
        & 0.936
        & \textbf{0.950}
        & 0.842 \\

        \rowcolor{oursblue}
        \hspace{0.5em}+\method (Ours)
        & \textbf{0.816}
        & \textbf{0.910}
        & \textbf{68.35}
        & \textbf{0.939}
        & 0.949
        & \textbf{0.861} \\

        \addlinespace[1pt]

        WorldMem
        & 0.782
        & 0.924
        & 70.23
        & 0.910
        & 0.916
        & 0.866 \\

        \rowcolor{oursblue}
        \hspace{0.5em}+\method (Ours)
        & \textbf{0.793}
        & \textbf{0.932}
        & \textbf{70.47}
        & \textbf{0.913}
        & \textbf{0.918}
        & \textbf{0.889} \\

        \addlinespace[1pt]

        YuMe 1.5
        & 0.765
        & 0.872
        & 50.98
        & 0.944
        & 0.965
        & 0.883 \\

        \rowcolor{oursblue}
        \hspace{0.5em}+\method (Ours)
        & \textbf{0.798}
        & \textbf{0.886}
        & \textbf{51.80}
        & \textbf{0.947}
        & \textbf{0.967}
        & \textbf{0.912} \\

        \bottomrule
    \end{tabular}}
\end{table}

\begin{figure}[t]
    \setlength{\abovecaptionskip}{6pt}
    \setlength{\belowcaptionskip}{0pt}
    \centering
    \includegraphics[width=\linewidth]{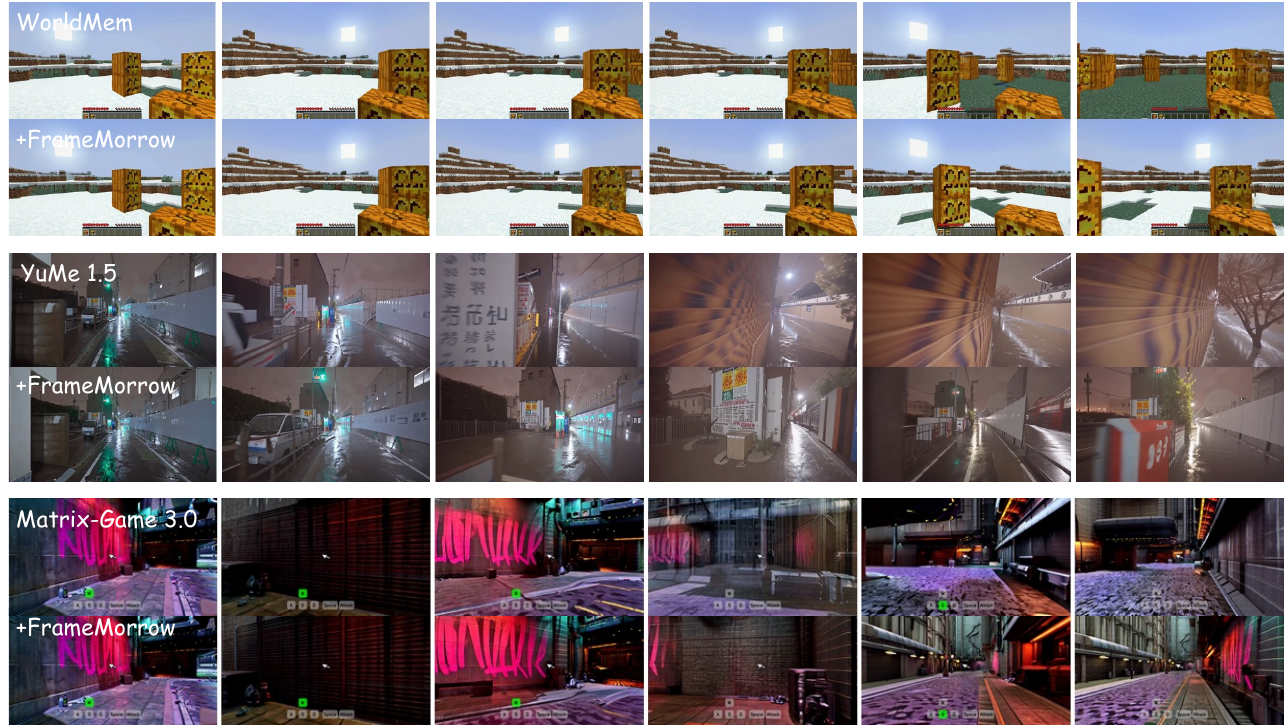}
    \vspace{-5mm}
    \caption{\textbf{Qualitative comparisons for interactive world generation.}
    Each pair shows the original backbone (top) and +\method (bottom)
    on WorldMem, YuMe 1.5, and Matrix-Game 3.0.}
    \label{fig:world_qualitative}
    \vspace{-1mm}
\end{figure}

\noindent
\textbf{Action-conditioned interactive world models.}
\label{sec:world_model_results}
Table~\ref{tab:world_models} shows consistent improvements across the
evaluated world models. \method improves subject and background consistency,
imaging quality, anti-flicker, and action alignment on all three backbones.
Action alignment increases by 0.019 for Matrix-Game 3.0, 0.023 for WorldMem,
and 0.029 for YuMe 1.5, while motion quality is largely preserved.
These results demonstrate that \method strengthens both long-range visual
consistency and control alignment across diverse interactive world models.
Figure~\ref{fig:world_qualitative} provides corresponding visual examples.
WorldMem better preserves the snow-covered ground, while YuMe 1.5 and
Matrix-Game 3.0 retain more consistent street and corridor structures over
longer rollouts.

\subsection{Analysis and Ablations}
\label{sec:ablation}

We analyze three design choices on the 60-second single-shot interactive
generation setting with a frozen Self-Forcing backbone. The recent-context
window retains the backbone's native default, and the eligible-history rule
is fixed. Only older historical frames count toward $K$.

\begin{figure}[!t]
    \centering
    \includegraphics[width=\textwidth]{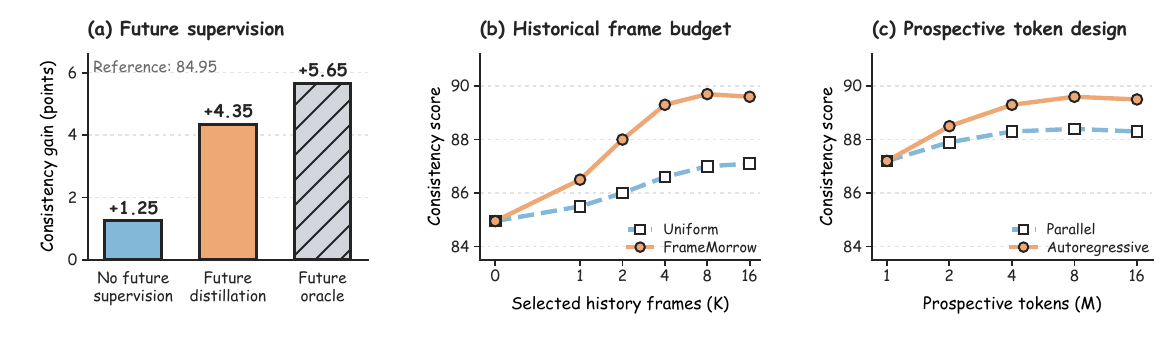}
    \vspace{-8mm}
    \caption{\textbf{Ablation studies.}
    Future supervision, frame budget, and token design.}
    \vspace{-2mm}
    \label{fig:ablation}
\end{figure}

\noindent
\textbf{Future supervision.}
Figure~\ref{fig:ablation}(a) compares recent-context supervision
(the no-future control), future distillation, and a future-informed oracle.
The no-future control constructs ranking targets from DINOv2 similarities
between eligible historical frames and recent-context observations,
whereas future distillation uses realized future observations. Their gains over the native score of 84.95 are 1.25, 4.35,
and 5.65 points. Future distillation adds 3.10 points over the control and
reduces the oracle gap from 4.40 to 1.30 points, supporting continuation-derived
targets for causal selection. The oracle accesses reference future observations and serves
as a diagnostic comparison.

\noindent
\textbf{Historical frame budget.}
Figure~\ref{fig:ablation}(b) varies the number of selected historical frames $K$
while fixing $M=4$, and compares \method with uniform sampling from the same eligible history.
\method reaches 88.00 with only two selected frames, exceeding uniform sampling
with sixteen frames (87.10). Increasing $K$ from 4 to 8 adds only 0.40 points,
with no further improvement at $K=16$.
These results show that selecting relevant historical frames matters more than simply increasing their number.

\noindent
\textbf{Prospective token design.}
Figure~\ref{fig:ablation}(c) compares autoregressive and parallel
prediction of prospective tokens at $K=4$.
Increasing the number of autoregressively predicted tokens from
$M=1$ to $M=4$ raises consistency from 87.20 to 89.30.
At $M=4$, autoregressive prediction exceeds parallel prediction
by 1.00 point.
Increasing $M$ to 8 adds only 0.30 points, with no further gain at $M=16$.
Four autoregressive tokens capture most of the improvement.

\begin{wraptable}[10]{r}{0.42\linewidth}
\vspace{-2mm}
    \centering
    \small
    \setlength{\abovecaptionskip}{0pt}
    \setlength{\belowcaptionskip}{3pt}
    \caption{\textbf{Historical frame selection.}}
    \label{tab:selection_strategies}
    \setlength{\tabcolsep}{3pt}
    \renewcommand{\arraystretch}{1.08}
    \begin{tabular}{@{}lrr@{}}
        \toprule
        \textbf{Strategy} & \textbf{Single-shot} & \textbf{Multi-shot} \\
        \midrule
        Recent-$K$ & 85.70 & 92.80 \\
        Uniform & 86.60 & 92.21 \\
        Context matching & 87.85 & 92.68 \\
        Prompt matching & 88.35 & 93.17 \\
        Direct scoring & 88.22 & 92.53 \\
        \rowcolor{oursblue}\textbf{\method} & \textbf{89.30} & \textbf{93.49} \\
        \bottomrule
    \end{tabular}
\end{wraptable}

\noindent
\textbf{Historical frame selection.}
Table~\ref{tab:selection_strategies} compares consistency for 60-second
single-shot generation with Self-Forcing and multi-shot generation with
LongLive 2.0. All strategies share the backbone, eligible history, memory
budget, and conditioning interface.
\textit{1) Recent-$K$} selects the latest eligible frames.
\textit{2) Uniform} samples frames uniformly from the eligible history.
\textit{3) Context matching} ranks frames by cosine similarity to the mean
recent-context feature from the selector's frozen visual encoder.
\textit{4) Prompt matching} ranks frames by frozen CLIP image--text similarity
to the next-shot or next-segment prompt.
\textit{5) Direct scoring} uses the same inputs to predict each candidate's score from its frame without prospective tokens.
\method outperforms the five selection strategies.

\begin{wraptable}[10]{r}{0.45\linewidth}
\vspace{-2mm}
    \centering
    \small
    \setlength{\abovecaptionskip}{0pt}
    \setlength{\belowcaptionskip}{3pt}
    \caption{\textbf{Inference overhead.}}
    \label{tab:inference_overhead}
    \setlength{\tabcolsep}{2.5pt}

    \begin{tabular}{@{}lrrr@{}}
        \toprule
        \textbf{Method}
        & \textbf{Sel. (ms)}
        & \textbf{Total (s)}
        & $\boldsymbol{\Delta}$ (\%) \\
        \midrule

        Self-Forcing
        & -- & 82.6 & -- \\
        \rowcolor{oursblue}
        \hspace{0.5em}+\method
        & 112 & 84.0 & 1.7 \\

        \addlinespace[1pt]

        LongLive 1.0
        & -- & 67.8 & -- \\
        \rowcolor{oursblue}
        \hspace{0.5em}+\method
        & 109 & 69.2 & 2.1 \\

        \addlinespace[1pt]

        Causal Forcing
        & -- & 82.6 & -- \\
        \rowcolor{oursblue}
        \hspace{0.5em}+\method
        & 115 & 84.0 & 1.7 \\

        \bottomrule
    \end{tabular}
\end{wraptable}

\noindent
\textbf{Inference overhead.}
We measure the inference overhead of \method on three models under matched hardware and sampling
settings (Table~\ref{tab:inference_overhead}).
For all backbones, \method uses $K=4$ and $M=4$, with selection
refreshed every 5 seconds.
Selection time includes feature encoding, prospective token prediction,
scoring, and Top-$K$ per refresh.
Total time covers the pipeline, including historical-frame
conditioning, and $\Delta$ denotes the increase over the native backbone.


%% file: sec/conclusion.tex
\section{Conclusion}
\label{sec:conclusion}

We present \method, which uses \emph{prospective tokens} to predict future information needs and select relevant historical frames.
Its explicit-frame interface enables lightweight, plug-and-play use across diverse generators.
Experiments on five benchmarks and 11 models demonstrate improvements in long-range consistency, visual quality, and action alignment.
These results support selecting historical information according to future needs rather than current relevance alone.

%% file: sec/appendix.tex
\newtcolorbox{paperprompt}[1]{
  breakable, colback=white, colframe=black!45,
  colbacktitle=black!5, coltitle=black,
  boxrule=0.5pt, arc=1mm,
  left=8pt, right=8pt, top=6pt, bottom=6pt,
  toptitle=5pt, bottomtitle=5pt,
  fonttitle=\small\sffamily\bfseries, fontupper=\small,
  before skip=8pt, after skip=8pt,
  title={#1}}

\section{Implementation Details}
\label{app:implementation}

\subsection{Selector Architecture}
\label{app:architecture}
The selector takes eligible history, recent context, and the known rollout condition. Visual and condition encoders are frozen. Only the visual/condition projections, causal Transformer, and ranking query/key projections are trained, without the generator. Compatible text-conditioned backbones share one checkpoint. Action-conditioned experiments use a separate checkpoint with the same architecture.

Table~\ref{tab:app_architecture} summarizes the architecture. DINOv2 ViT-B/14 provides final class-token frame features for the visual encoder and teacher. Learned projections map visual and condition features into the selector space. Text conditions use frozen UMT5-XXL token features. Actions are serialized as short motion descriptions and encoded by frozen CLIP ViT-B/32, with a separate projection for its feature dimension.

Historical and recent tokens follow observation time, with learned temporal and token-type embeddings distinguishing history, recent context, and conditions. A learned start token initializes autoregressive queries under a causal mask. Padding is masked in attention and ranking. Only eligible history is scored and returned by Top-$K$, excluding recent context. Deployment retains the backbone's native recent window. If fewer than $K$ candidates exist, all are returned without duplication.

\begin{table}[ht]
\centering
\caption{\textbf{Selector architecture.} Frozen encoders are excluded from the trainable parameter estimate.}
\label{tab:app_architecture}
\begin{tabular}{ll}
\toprule
Configuration & Value \\
\midrule
Visual encoder / teacher & DINOv2 ViT-B/14 \\
Visual feature dimension & 768 \\
Transformer layers / hidden dimension & 4 / 512 \\
Attention heads / FFN dimension & 8 / 2,048 \\
Normalization / activation & Pre-LayerNorm / GELU \\
Dropout & 0.0 \\
Positional encoding & Learned temporal embeddings \\
Prospective queries $M$ / memory budget $K$ & 4 / 4 \\
Maximum training candidates & 128 \\
Trainable parameters, text selector &  16M \\
\bottomrule
\end{tabular}
\end{table}

\subsection{Ranking Targets and Losses}
\label{app:ranking_details}
The teacher computes cosine similarities between $\ell_2$-normalized DINOv2 features of historical and realized future frames. Future observations supply detached targets only and never enter the selector. Scores are aggregated by the smooth maximum in Section~\ref{sec:future_supervision}, then softmaxed over valid historical candidates for listwise supervision, without per-example min--max normalization.

The teacher uses $H=8$ continuation frames sampled at two fps over the next four seconds. Table~\ref{tab:app_loss} lists the loss settings. Pairwise softplus loss is averaged over all ordered pairs with teacher-score gaps of at least $\delta$, or set to zero if none qualify. The listwise term remains active. Both losses are averaged across examples, not pooled candidates.

\begin{table}[t]
\centering
\caption{\textbf{Ranking hyperparameters.}}
\label{tab:app_loss}
\begin{tabular}{lll}
\toprule
Symbol & Role & Value \\
\midrule
$\tau_q$ & Query-score aggregation & 0.10 \\
$\tau_f$ & Future-similarity aggregation & 0.05 \\
$\tau_r$ & Teacher ranking temperature & 0.10 \\
$\tau_s$ & Student ranking temperature & 1.00 \\
$\delta$ & Teacher margin for pair inclusion & 0.05 \\
$\lambda$ & Pairwise loss weight & 0.50 \\
$H$ & Number of teacher continuation frames & 8 \\
\bottomrule
\end{tabular}
\end{table}

\subsection{Optimization and Compute}
\label{app:optimization}
Text- and action-conditioned selectors are trained independently using precomputed frozen encoder features. Table~\ref{tab:app_training} gives the configuration. Linear warmup precedes cosine learning-rate decay, gradients are clipped before each update, and validation listwise loss selects the final checkpoint. The planning budget is six hours on two A100 80GB GPUs, or 12 GPU-hours per selector, excluding annotation and feature extraction.

\begin{table}[t]
\centering
\caption{\textbf{Optimization settings.}}
\label{tab:app_training}
\begin{tabular}{ll}
\toprule
Configuration & Value \\
\midrule
Optimizer & AdamW \\
Learning rate / weight decay & $10^{-4}$ / 0.01 \\
Adam coefficients & $(0.9,0.999)$ \\
Schedule / warmup & Cosine / 5\% of updates \\
Global batch size / epochs & 64 / 5 \\
Precision / gradient clipping & BF16 / global norm 1.0 \\
Training seed & 0 \\
Hardware budget & $2\times$ A100 80GB \\
\bottomrule
\end{tabular}
\end{table}

\section{Training Data Construction}
\label{app:data_construction}
\subsection{Text-conditioned Training Tuples}
We use 10K OpenVidHD videos, divided into two-second annotation units with four uniformly sampled frames each. These units do not set rollout-condition durations or selection refresh intervals. Qwen3-VL-8B-Instruct generates structured segment descriptions, which form the streaming conditions.

Each tuple $(\mathcal{H}_t,\mathcal{L}_t,c_t^+,\mathcal{Y}_t^+)$ contains eligible history, recent context ending at $t$, the next-rollout condition, and a teacher-only continuation. Recent context comprises eight frames from the preceding four seconds. Up to 128 historical frames are sampled uniformly before this window and kept in temporal order. Boundaries lacking eligible history or a complete continuation are skipped. This training window does not replace the backbone's native deployment window.

Videos are split 90:10 for training and validation before tuple extraction, keeping all tuples from each video together. Teacher frames remain within the condition's temporal scope. Shorter intervals use fewer frames rather than sampling beyond that scope.

\subsection{Segment Annotation and Prompt Construction}
\label{app:annotation_prompt}
The annotation template preserves visible identities, scene attributes, actions, and state changes without introducing unsupported objects or events.

\begin{paperprompt}{Prompt 1: Segment annotation}
\textbf{Instructions}\par
Given four chronologically ordered frames from a two-second video segment, describe the visible subjects, scene, actions, and object states. If a preceding segment description is supplied, identify visible changes while preserving consistent names for recurring subjects and objects. Do not infer unseen events or future actions.

\smallskip
\textbf{Output}\par
Return the fields \texttt{subject}, \texttt{scene}, \texttt{action}, \texttt{state}, \texttt{change}, and \texttt{segment\_prompt}. The \texttt{segment\_prompt} should be a concise visual description suitable for conditioning this segment.
\end{paperprompt}

Structured fields form each segment prompt, excluding later timestamps and teacher scores. Annotation uses temperature zero and at most 256 output tokens. Empty or malformed outputs are regenerated once, then excluded if still invalid.

\subsection{Action-conditioned Training Tuples}
We use 10K chunk-aligned Sekai Game-Walking examples. Each combines observed history and context, the next chunk's pseudo-action condition, and its realized continuation for the teacher. Pseudo-actions derive from relative camera translation and rotation and need not reproduce the original keyboard inputs.

Under the camera-to-world pose convention, let $R_t$ and $\mathbf{p}_t$ denote camera orientation and position. Local translation and angular velocity are computed as
\begin{equation}
\mathbf{v}_t=R_t^\top(\mathbf{p}_{t+1}-\mathbf{p}_t)/\Delta t,
\qquad
\boldsymbol{\omega}_t=\operatorname{Log}(R_t^\top R_{t+1})^\vee/\Delta t.
\label{eq:app_camera_motion}
\end{equation}
Translation is normalized by the trajectory's median nonzero speed, with components activated above 0.2. Yaw and pitch use a $2^\circ$/s threshold. Labels cover forward/backward, left/right, turn left/right, and look up/down. Simultaneous labels are retained, and chunks are stationary when all components fall below threshold. A trajectory-level 90:10 train/validation split keeps adjacent chunks together.

\section{Backbone Integration Details}
\label{app:backbone_integration}
\subsection{Shared Selection and Backbone-specific Conditioning}
The selector returns explicit frames and shares one checkpoint per condition modality across compatible generators, with $K=M=4$. Each backbone retains its native recent window. Integration trains neither the generator nor a new conditioning module.

Selected frames are sorted by timestamp and converted by each backbone's adapter into its reference representation (Table~\ref{tab:app_integration}). Generator activations never feed back into the selector.

\begin{table}[t]
\centering
\caption{\textbf{Backbone integration map.} Every row retains the native recent window and selects up to four additional historical frames. No generator fine-tuning is used.}
\label{tab:app_integration}
\small
\begin{tabular}{@{}p{0.24\linewidth}p{0.14\linewidth}p{0.55\linewidth}@{}}
\toprule
Backbone & Selector & Historical conditioning route \\
\midrule
Self-Forcing & Text & Frame encoding followed by historical KV conditioning \\
LongLive 1.0 & Text & Historical frame encoding into the memory cache \\
Causal Forcing & Text & Frame encoding into causal context KV states \\
CausVid & Text & Historical latent encoding into context KV states \\
LongLive 2.0 & Text & Historical references in the streaming memory interface \\
ShotStream & Text & Selected images as cross-shot reference conditioning \\
Matrix-Game 3.0 & Action & Encoded historical frames in the rollout context \\
WorldMem & Action & Historical frame memory with associated pose metadata \\
YuMe 1.5 & Action & Historical latent references for rollout conditioning \\
\bottomrule
\end{tabular}
\end{table}

\subsection{Temporal Positions and Cache Management}
For KV conditioning, the backbone's frozen image/video encoder and context-encoding path convert selected frames into layer-specific keys and values. DINOv2 features are not copied into this cache. Reference latents are conditioning inputs, not generated output frames. Backbones with direct image-reference interfaces instead receive selected RGB frames.

Selection retains original timestamps. Adapters use historical temporal positions where supported, or map sorted frames to reference slots in temporal order. Refreshes replace historical slots while preserving native recent context, whose frames are excluded from the candidate pool. Thus, the four-frame budget fixes additional observations, not latent-token or KV-cache size across backbones.

\subsection{Matched Integration Controls}
Recent-$K$, Uniform, Context matching, Prompt matching, and \method share eligible history, budget, and conditioning adapter within each backbone. Context matching uses the selector's recent visual features, whereas Prompt matching uses CLIP image--text similarity to the known condition. These controls isolate selection quality. Comparisons with the unaugmented backbone also include the benefit of adding historical references.

\section{Evaluation Protocols}
\label{app:evaluation}

Unless otherwise specified, historical frame selection is refreshed every five seconds of generated video. Each refresh uses only the eligible history, recent context, and condition available at that time.
\subsection{Long Video Generation}
We generate 60-second videos from 128 MovieGenBench prompts and evaluate six VBench-Long dimensions: subject/background consistency, motion smoothness, dynamic degree, aesthetic quality, and imaging quality. They assess foreground/scene persistence, temporal smoothness, motion magnitude, aesthetics, and frame-level quality, respectively. Higher is better for all six.

Prompt $i\in\{0,\ldots,127\}$ uses paired seed $i$. Each backbone and its augmented variants share its released resolution, frame rate, denoising steps, and guidance. Settings are matched within, not across, backbones. Evaluation covers the full 60 seconds without selecting favorable subsequences.

Methods are ranked per metric within each backbone before display rounding, with ties assigned average ranks. Average rank is the mean over six metrics. Dataset scores weight prompts equally.

\subsection{Interactive Video Generation}
\label{app:interactive_protocol}
We evaluate 100 single-shot and 100 multi-shot cases, each comprising six ten-second condition intervals. LLM-generated prompt sequences require dependencies across intervals. Single-shot cases retain one scene, while multi-shot transitions depend on previously established subjects and states. Methods share prompts and recent context. Selection refreshes every five seconds, independently of the ten-second condition/evaluation intervals.

Segment-wise alignment pairs each ten-second interval with its current prompt, following the evaluation organization used for LongLive. In the CLIP ViT-B/32 implementation, eight uniformly sampled frames per interval are encoded and compared with the normalized prompt embedding. The score for interval $j$ is
\begin{equation}
s_j=\frac{100}{n_j}\sum_{u=1}^{n_j}
\cos\bigl(E_{\mathrm{CLIP}}^{\mathrm{img}}(x_{j,u}),
E_{\mathrm{CLIP}}^{\mathrm{txt}}(c_j)\bigr).
\label{eq:app_segment_clip}
\end{equation}
Each interval's score is averaged across cases using only its current prompt, never prompts from later intervals.

Quality, Consistency, and Aesthetic use the VBench quality aggregate, subject consistency, and aesthetic evaluator, respectively. Multi-shot evaluation measures recurring-subject consistency across shots and temporal quality within shots, avoiding penalties for intended scene changes.

\subsection{Closed-Source Generation}

We evaluate Seedance~2.0 and Kling~O3 through their public reference-conditioning interfaces via fal.ai on 30 cases with five turns each. Methods share prompts, initial context, explicit API seeds, and reference budgets. The native model receives the latest clip as a video reference. Uniform-$K$ and \method additionally receive equal numbers of historical image references, differing only in frame selection. Other settings remain fixed within each backbone.

\subsection{Action-conditioned World Models}
\label{app:action_alignment}
We evaluate Matrix-Game~3.0, WorldMem, and YuMe~1.5 on 60 trajectories each. Native and \method variants share initial context and actions. Visual metrics cover subject/background consistency, anti-flicker, motion smoothness, and imaging quality. Qwen3-VL-8B-Instruct judges whether rollouts follow the supplied controls.

The judge receives 16 ordered frames per action interval, command descriptions, and durations, without method identities or competing outputs. Each interval is scored once at temperature zero using Table~\ref{tab:app_action_rubric}. Scores are averaged within trajectories and then across trajectories, avoiding extra weight for longer sequences.

\begin{table}[t]
\centering
\caption{\textbf{Action-alignment rubric.}}
\label{tab:app_action_rubric}
\begin{tabular}{@{}cp{0.84\linewidth}@{}}
\toprule
Score & Criterion \\
\midrule
1.00 & All commanded motion directions and their temporal order are clearly followed. \\
0.75 & The dominant commands are followed, with minor delay or one brief inconsistency. \\
0.50 & Some commands are followed, but a substantial part is missing or ambiguous. \\
0.25 & Only weak evidence supports the commands, with mostly inconsistent motion. \\
0.00 & The commanded motion is absent, reversed, or unsupported by the visible rollout. \\
\bottomrule
\end{tabular}
\end{table}

\begin{paperprompt}{Prompt 2: Action-alignment evaluation}
\textbf{Instructions}\par
Compare camera motion with the supplied controls and intervals. Distinguish camera motion from moving objects. Judge action compliance, not aesthetics. Using the rubric, assign 0, 0.25, 0.5, 0.75, or 1. Do not infer invisible motion.

\smallskip
\textbf{Inputs}\par
Controls: \texttt{<action sequence and intervals>}\par
Frames: \texttt{<chronologically sampled rollout frames>}

\smallskip
\textbf{Output}\par
Return JSON fields \texttt{score} and \texttt{reason}, citing visible evidence.
\end{paperprompt}

Malformed outputs are retried once, with persistent failures reported as missing. The judge assesses observable action compliance rather than physical correctness.

\section{Additional Ablations}
\label{app:additional_ablations}

Unless otherwise specified, additional ablations use 60-second single-shot interactive generation with frozen Self-Forcing, its native recent window, $K=4$ historical frames, and $M=4$ prospective queries. We report overall consistency, as in the main ablations.


\subsection{Sensitivity to the Visual Teacher}
\label{app:teacher_encoder}

\paragraph{Visual teacher.}
We replace the DINOv2 teacher with CLIP or SigLIP, fixing selector architecture, training trajectories, candidate history, and optimization. All three improve consistency (Table~\ref{tab:teacher_encoder}), supporting the use of different visual feature spaces. DINOv2 performs best and is used in the main experiments.

\begin{table}[t]
    \centering
    \caption{\textbf{Sensitivity to the visual teacher.}
    All variants use the same selector architecture and training data.
    $\Delta$ denotes the improvement over the native Self-Forcing
    consistency score of 84.95.}
    \label{tab:teacher_encoder}

    \setlength{\tabcolsep}{9pt}
    \renewcommand{\arraystretch}{1.05}
    \begin{tabular}{lcc}
        \toprule
        \textbf{Visual Teacher}
        & \textbf{Consistency $\uparrow$}
        & $\boldsymbol{\Delta}$ \\
        \midrule

        CLIP
        & 88.42
        & +3.47 \\

        SigLIP
        & 88.76
        & +3.81 \\

        \rowcolor{oursblue}
        DINOv2
        & \textbf{89.30}
        & \textbf{+4.35} \\

        \bottomrule
    \end{tabular}
\end{table}

\subsection{Ranking Objective}
\label{app:ranking_objective}

\paragraph{Listwise and pairwise supervision.}
Listwise loss transfers the teacher's relevance distribution, while pairwise loss preserves orderings above the teacher-score margin. Table~\ref{tab:ranking_loss} compares each alone with their combination. Both help, listwise supervision is stronger alone, and adding pairwise constraints yields the best consistency.

\begin{table}[t]
    \centering
    \caption{\textbf{Ablation of the ranking objective.}
    The full objective combines listwise distribution matching with
    margin-based pairwise ordering constraints.}
    \label{tab:ranking_loss}

    \setlength{\tabcolsep}{9pt}
    \renewcommand{\arraystretch}{1.05}
    \begin{tabular}{lcc}
        \toprule
        \textbf{Objective}
        & \textbf{Consistency $\uparrow$}
        & $\boldsymbol{\Delta}$ \\
        \midrule

        Pairwise only
        & 88.21
        & +3.26 \\

        Listwise only
        & 88.67
        & +3.72 \\

        \rowcolor{oursblue}
        Listwise + Pairwise
        & \textbf{89.30}
        & \textbf{+4.35} \\

        \bottomrule
    \end{tabular}
\end{table}

\subsection{Length of the Realized Continuation}
\label{app:future_horizon}

\paragraph{Continuation horizon.}
We vary the number of teacher continuation observations $H$, which affects supervision but not deployed inputs or computation. Consistency improves substantially up to $H=4$, then largely saturates (Table~\ref{tab:future_horizon}). Multiple future observations provide a more reliable target than one frame, with limited benefit from longer horizons.

\begin{table}[t]
    \centering
    \caption{\textbf{Effect of the realized-continuation horizon.}
    $H$ denotes the number of continuation observations available only to
    the training teacher.}
    \label{tab:future_horizon}

    \setlength{\tabcolsep}{10pt}
    \renewcommand{\arraystretch}{1.05}
    \begin{tabular}{ccc}
        \toprule
        $\boldsymbol{H}$
        & \textbf{Consistency $\uparrow$}
        & $\boldsymbol{\Delta}$ \\
        \midrule

        1
        & 87.86
        & +2.91 \\

        2
        & 88.71
        & +3.76 \\

        4
        & 89.14
        & +4.19 \\

        \rowcolor{oursblue}
        8
        & \textbf{89.30}
        & \textbf{+4.35} \\

        \bottomrule
    \end{tabular}
\end{table}

\subsection{Aggregation of Prospective Queries}
\label{app:query_aggregation}

\paragraph{Query-score aggregation.}
We compare mean, hard-maximum, and smooth-maximum pooling of query--frame scores. Mean pooling favors consistent matches across queries, while hard maximum keeps only the strongest match. Smooth maximum allows one strong prospective match to dominate while retaining other queries' contributions and gives the best consistency (Table~\ref{tab:query_aggregation}).

\begin{table}[t]
    \centering
    \caption{\textbf{Aggregation across prospective queries.}
    All variants use $M=4$ queries and the same trained selector
    configuration.}
    \label{tab:query_aggregation}

    \setlength{\tabcolsep}{10pt}
    \renewcommand{\arraystretch}{1.05}
    \begin{tabular}{lcc}
        \toprule
        \textbf{Aggregation}
        & \textbf{Consistency $\uparrow$}
        & $\boldsymbol{\Delta}$ \\
        \midrule

        Mean
        & 88.43
        & +3.48 \\

        Hard maximum
        & 88.91
        & +3.96 \\

        \rowcolor{oursblue}
        Smooth maximum
        & \textbf{89.30}
        & \textbf{+4.35} \\

        \bottomrule
    \end{tabular}
\end{table}

\subsection{Relation Between Future Relevance and Generation Utility}
\label{app:utility_validation}

\paragraph{Generation-utility validation.}
The target in Eq.~(5) measures visual correspondence with the realized continuation, which need not imply downstream benefit. We test whether teacher relevance $r^F_{t,i}$ correlates with the generation utility of individual historical frames.

This diagnostic uses condition-free long-video generation. All ranking signals exclude upcoming conditions. At each rollout step $t$, we sample $N_u$ eligible candidates $x_{\tau_i}\in\mathcal{H}_t$.

The privileged teacher score $r^F_{t,i}$ uses the held-out realized continuation, not generated futures. Generated continuations are used only to measure each candidate's downstream utility.

We first generate a baseline continuation using only the backbone's native recent
context,
\begin{equation}
    \hat{\mathcal{Y}}^{\,0}_t
    =
    G_{\omega}(\mathcal{L}_t),
    \label{eq:utility_base}
\end{equation}
and then generate an additional continuation for each candidate while providing
that candidate through the same historical-conditioning interface,
\begin{equation}
    \hat{\mathcal{Y}}^{\,i}_t
    =
    G_{\omega}(\mathcal{L}_t, x_{\tau_i}).
    \label{eq:utility_candidate}
\end{equation}
All generations share the initial context, seed, sampling settings, and frozen backbone. Only the supplied historical frame differs.

We define the empirical generation utility of candidate $i$ as
\begin{equation}
    u_{t,i}
    =
    Q\!\left(\hat{\mathcal{Y}}^{\,i}_t\right)
    -
    Q\!\left(\hat{\mathcal{Y}}^{\,0}_t\right),
    \label{eq:generation_utility}
\end{equation}
where $Q$ is the main experiments' long-range consistency evaluator. It measures preservation of subjects, objects, and scene appearance from preceding context, rather than internal consistency of the new continuation alone. Positive utility indicates better preservation than the native backbone.

Because trajectory difficulty and utility scale may vary across examples, we
measure rank correspondence among historical candidates at each rollout step
rather than pooling raw utility values globally. Specifically, we compute
\begin{equation}
    \rho_t(s,u)
    =
    \operatorname{Spearman}
    \left(
        \{s_{t,i}\}_{i=1}^{N_u},
        \{u_{t,i}\}_{i=1}^{N_u}
    \right),
    \label{eq:utility_spearman}
\end{equation}
where $s$ denotes a historical-frame ranking signal. We compare current-context
similarity, the privileged future-grounded teacher score $r^F$, and the deployed
\method prediction $\hat{r}$.

We average valid step-level correlations within each trajectory, then across trajectories. The analysis includes $N_{\mathrm{traj}}=\mathrm{30}$ held-out trajectories, $N_{\mathrm{step}}=\mathrm{120}$ valid steps, and $N_u=\mathrm{8}$ candidates per step. We obtain 95\% confidence intervals by resampling trajectories with replacement 1,000 times.

\begin{table}[t]
    \centering
    \caption{\textbf{Relationship between historical-frame ranking signals and
    empirical generation utility.}
    Spearman correlations are computed across historical candidates at each
    rollout step, averaged within each trajectory, and then averaged across the
    held-out evaluation set. The future-grounded teacher uses the held-out realized
    continuation only for this diagnostic analysis, whereas \method does not
    observe future content at inference. Confidence intervals are obtained by
    trajectory-level bootstrap with 1,000 resamples.}
    \label{tab:utility_validation}

    \setlength{\tabcolsep}{7pt}
    \renewcommand{\arraystretch}{1.05}
    \begin{tabular}{lccc}
        \toprule
        \textbf{Ranking signal}
        & \textbf{Realized future}
        & $\boldsymbol{\rho}\boldsymbol{\uparrow}$
        & \textbf{95\% CI} \\
        \midrule

        Context similarity
        & No
        & 0.10
        & [0.04, 0.16] \\

        \method prediction $\hat{r}$
        & No
        & 0.42
        & [0.35, 0.49] \\

        Future-grounded teacher $r^F$
        & Yes
        & 0.51
        & [0.44, 0.57] \\

        \bottomrule
    \end{tabular}
\end{table}

Table~\ref{tab:utility_validation} shows weak correspondence between current-context similarity and generation utility. The teacher ranking aligns more closely with utility, and \method recovers much of this association without future access at inference. These results support continuation-based visual correspondence as a training proxy for downstream usefulness.

\section{Additional Qualitative Results}
\label{app:qualitative}
\begin{figure}[ht]
    \centering
    \includegraphics[width=\textwidth]
    {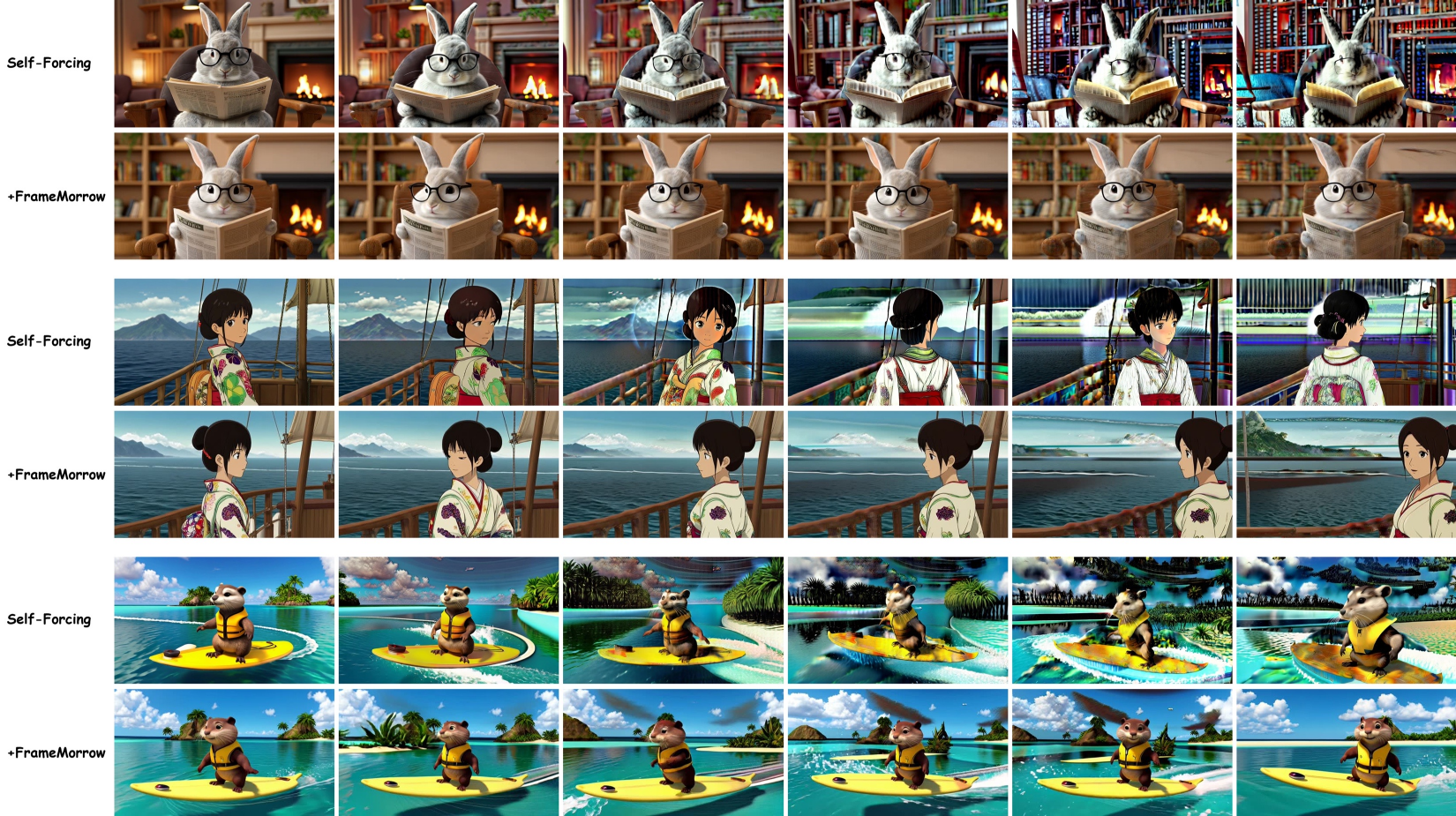}
    \caption{
    \textbf{Additional qualitative comparisons with Self-Forcing.}
    Each example compares the original Self-Forcing model (top)
    with Self-Forcing+\method (bottom) over long-horizon generation.
    \method better preserves subject appearance and visual details
    as generation progresses.
    }
    \label{fig:appendix_self_forcing}
\end{figure}

\begin{figure}[t]
    \centering
    \includegraphics[width=\textwidth]
    {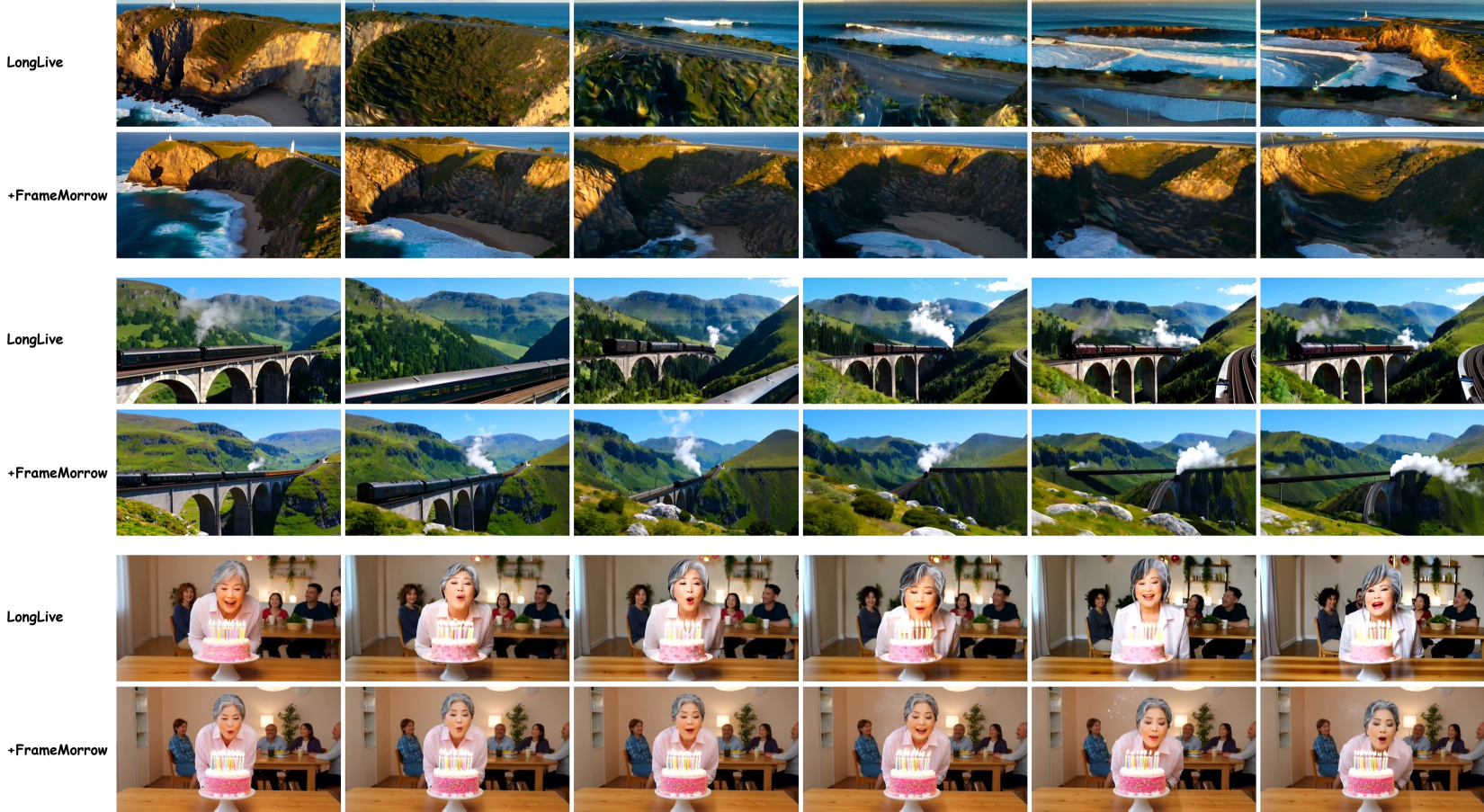}
    \caption{
    \textbf{Additional qualitative comparisons with LongLive.}
    Each example compares the original LongLive model (top)
    with LongLive+\method (bottom) under the same generation conditions.
    \method improves the preservation of subjects and scene appearance
    over extended generation.
    }
    \label{fig:appendix_longlive}
\end{figure}

\begin{figure}[t]
    \centering
    \includegraphics[width=\textwidth]
    {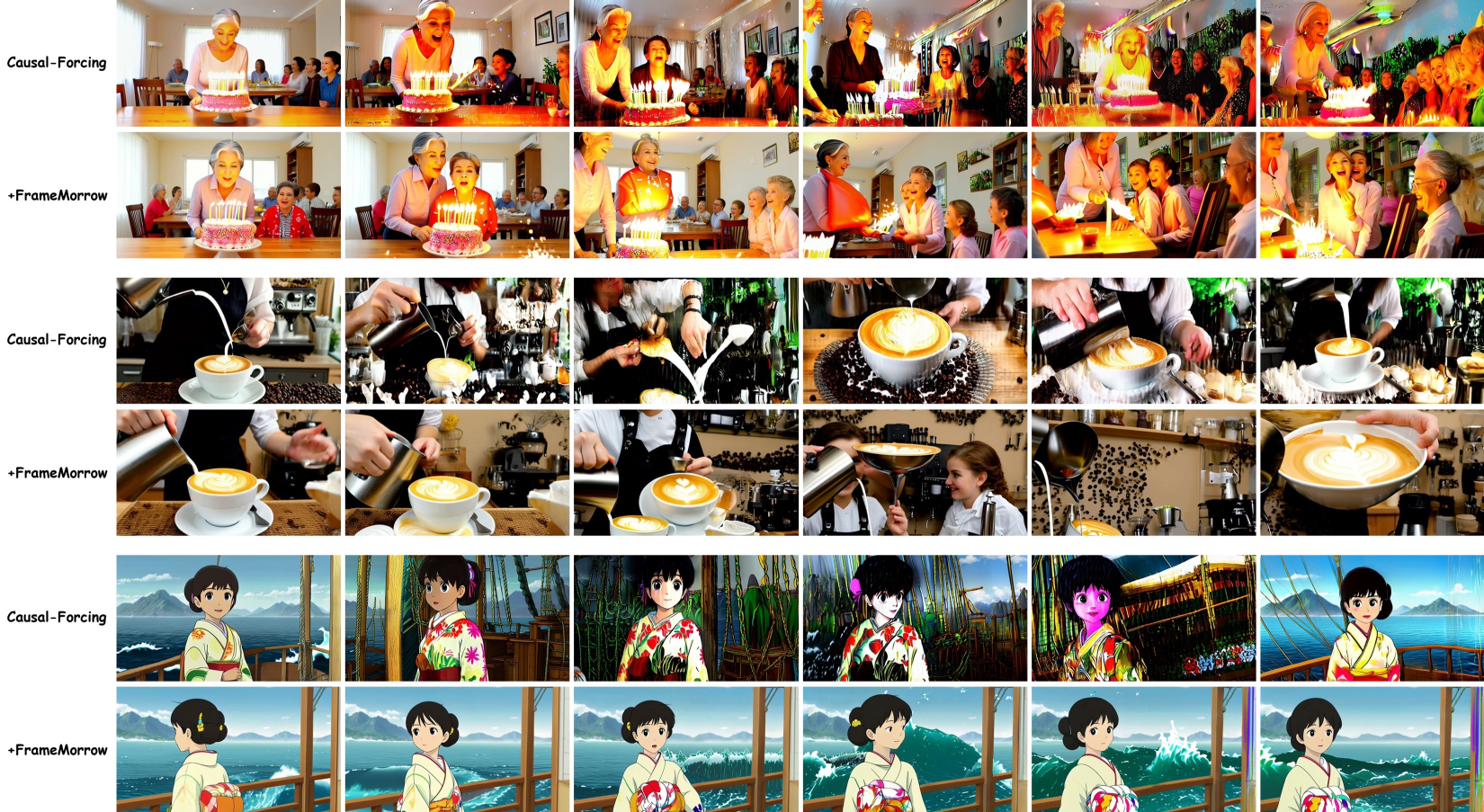}
    \caption{
    \textbf{Additional qualitative comparisons with Causal Forcing.}
    Each example compares the original Causal Forcing model (top)
    with Causal Forcing+\method (bottom) under the same generation conditions.
    \method provides more consistent subjects, objects, and scene details
    throughout long-horizon generation.
    }
    \label{fig:appendix_causal_forcing}
\end{figure}

Figures~\ref{fig:appendix_self_forcing},~\ref{fig:appendix_longlive}, and~\ref{fig:appendix_causal_forcing} compare native and \method-augmented Self-Forcing, LongLive, and Causal Forcing under matched generation conditions and sampling settings, illustrating long-horizon visual consistency.

\section{User Study}
\label{app:user_study}

We conduct a blinded pairwise study of long-video generation. For each of Self-Forcing, LongLive, and Causal Forcing, 30 randomly sampled native/\method video pairs share prompts and generation settings. All ten raters evaluate every pair, yielding 300 judgments per backbone and 900 per criterion. Method identities are hidden and left--right order is randomized. Raters independently assess \textit{long-range consistency} (preservation of subjects, objects, and scenes) and \textit{overall visual quality} (clarity, artifacts, and naturalness), selecting a preferred video or a tie when neither is clearly better.

\begin{figure}[t]
    \centering
    \includegraphics[width=\textwidth]{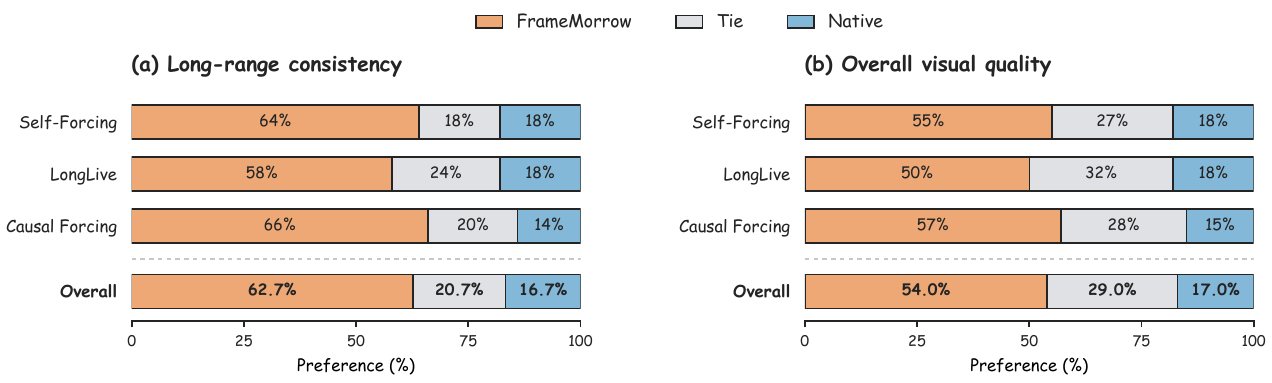}
    \caption{\textbf{User preferences for long-video generation.} \method is preferred over the native model across all three backbones. Overall averages the three backbone-level percentages.}
    \label{fig:user_study}
\end{figure}

Figure~\ref{fig:user_study} reports preferences, with Overall averaging the three backbones. \method is preferred more often than native models for both criteria on every backbone. Average consistency preferences are 62.7\% for \method, 16.7\% for native, and 20.7\% ties. Quality preferences are 54.0\%, 17.0\%, and 29.0\%, respectively. The stronger consistency preference matches our focus on historical preservation, while quality preferences indicate broader perceptual benefits.

